\documentclass[conference]{IEEEtran}
\IEEEoverridecommandlockouts

\usepackage{tikz}
\usepackage{float}
\usepackage{forest}
\usetikzlibrary{arrows.meta, shapes.geometric, positioning, fit, backgrounds}
\usepackage{booktabs}

\usepackage{cite}
\usepackage{amsmath,amssymb,amsfonts}
\usepackage{algorithmic}
\usepackage{graphicx}
\usepackage{textcomp}
\usepackage{xcolor}
\def\BibTeX{{\rm B\kern-.05em{\sc i\kern-.025em b}\kern-.08em
    T\kern-.1667em\lower.7ex\hbox{E}\kern-.125emX}}
\begin{document}

\title{Multimodal Representation Learning and Fusion
\thanks{\hspace*{-\parindent}\rule{3.8cm}{0.4pt} \\ 
$\ast$: Equal contribution. \\ 
$\dagger$: Corresponding author.}
}

\author{
    \IEEEauthorblockN{
        Qihang Jin$^{\ast}$\textsuperscript{1},
        Enze Ge$^{\ast}$\textsuperscript{2},
        Yuhang Xie$^{\ast}$\textsuperscript{1},
        Hongying Luo$^{\ast}$\textsuperscript{1},
        Junhao Song\textsuperscript{3}, \\
        Ziqian Bi\textsuperscript{3},
        Chia Xin Liang\textsuperscript{1},
        Jibin Guan\textsuperscript{4},
        Joe Yeong\textsuperscript{5},
        Xinyuan Song\textsuperscript{6},
        Junfeng Hao$^{\dagger}$\textsuperscript{4}
    }
    \\
    \IEEEauthorblockA{\textsuperscript{1}AI Agent Lab, Vokram Group, United Kingdom, ai-agent-lab@vokram.com}
    \IEEEauthorblockA{\textsuperscript{2}University of Bologna, Italy, enze.ge@studio.unibo.it}
    \IEEEauthorblockA{\textsuperscript{3}Vokram Group, United Kingdom, \{js, bizi\}@vokram.com}
    \IEEEauthorblockA{\textsuperscript{4}University of Minnesota, United States, jguan@umn.edu, ygzhjf85@gmail.com}
    \IEEEauthorblockA{\textsuperscript{5}Singapore General Hospital, Singapore, yeongps@imcb.a-star.edu.sg}
    \IEEEauthorblockA{\textsuperscript{6}Emory University, United States, xinyuan.song@emory.edu}
}


\maketitle

\begin{abstract}
Multi-modal learning is a fast-growing area in artificial intelligence. It tries to help machines understand complex things by combining information from different sources, like images, text, and audio. By using the strengths of each modality, multi-modal learning allows AI systems to build stronger and richer internal representations. These help machines better interpretation, reasoning, and making decisions in real-life situations. This field includes core techniques such as \textbf{representation learning} (to get shared features from different data types), \textbf{alignment methods} (to match information across modalities), and \textbf{fusion strategies} (to combine them by deep learning models). Although there has been good progress, some major problems still remain. Like dealing with different data formats, missing or incomplete inputs, and defending against adversarial attacks. Researchers now are exploring new methods, such as \textbf{unsupervised} or \textbf{semi-supervised learning}, \textbf{AutoML} tools, to make models more efficient and easier to scale. And also more attention on designing better evaluation metrics or building shared benchmarks, make it easier to compare model performance across tasks and domains. As the field continues to grow, multi-modal learning is expected to improve many areas: computer vision, natural language processing, speech recognition, and healthcare. In the future, it may help to build AI systems that can understand the world in a way more like humans, flexible, context-aware, and able to deal with real-world complexity.
 
\end{abstract}


\section{Introduction}

Multi-modal learning, a fast-growing research field that aims to enable machines to interpret and understand the world by combining different types of data, such as images, text, and audio \cite{2103.06304} \cite{1705.09406}. It tries to use the strengths of each modality to build a more complete and detailed understanding of complex things \cite{2301.04856}. This kind of learning is naturally cross-disciplinary, combining knowledge from computer vision, natural language processing, speech analysis, and other fields to make machines behave more like humans in perception and reasoning. 

One core concept in multi-modal learning is how we define ``multi-modality.'' Some recent works suggest using task-specific definitions \cite{2103.06304}, which focus on the features and representations that are important for each task. Different tasks have different needs.This flexible view helps to guide model design more precisely. Another important topic is building taxonomies to structure the field and classify methods based on how they work and what they are used for \cite{1705.09406}.

\begin{figure}[htbp]
    \centering
    \includegraphics[width=0.5\textwidth]{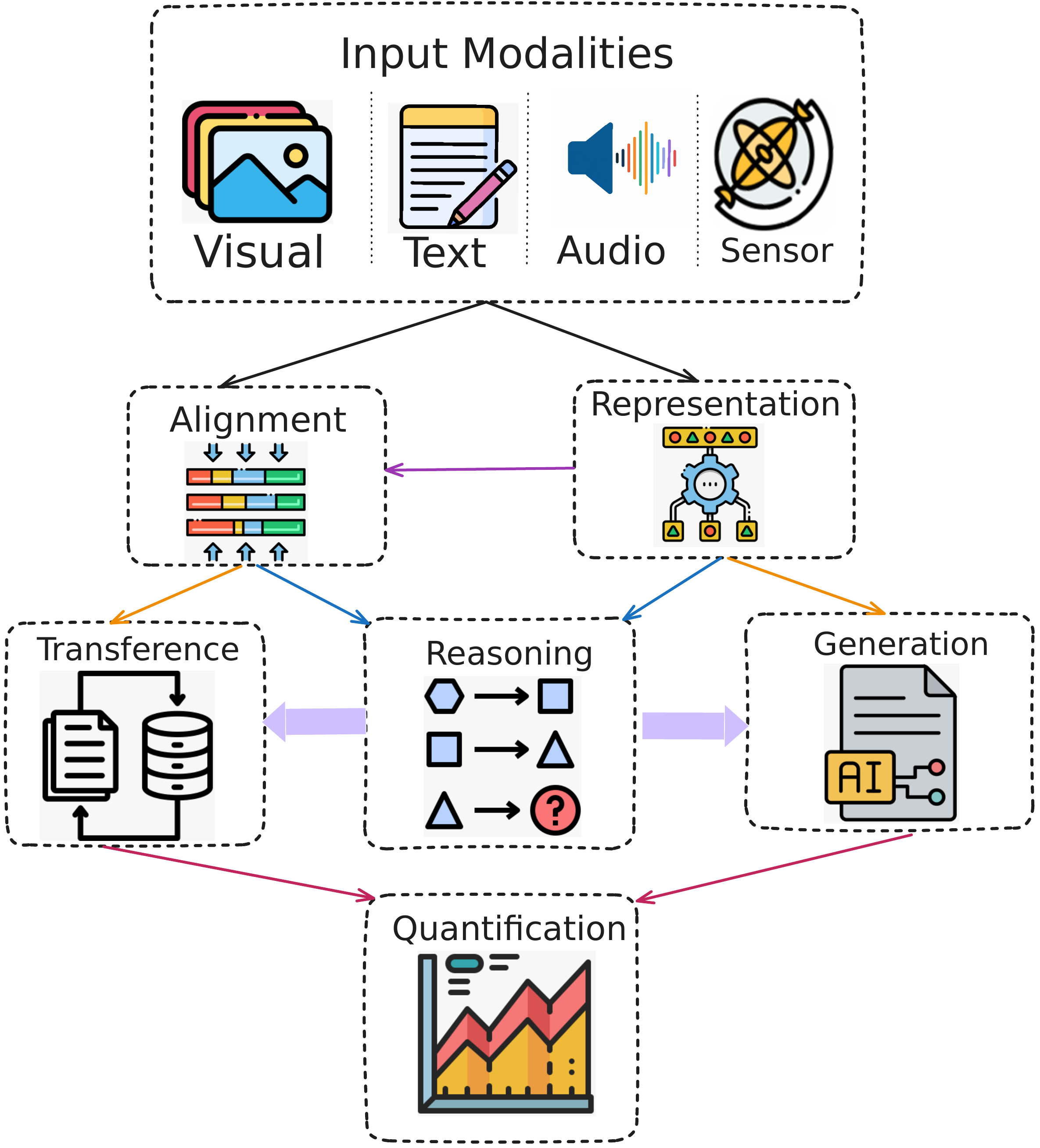}
    \caption{The core technical challenges in multi-modal learning. It begins with diverse Input Modalities (Visual, Text, Audio, Sensor) that are processed through Representation and Alignment. These foundational steps then support subsequent complex tasks including Reasoning, Generation, and Transference, while Quantification provides analytical evaluation across the framework.}
    \label{fig:multi-modal_intro}
\end{figure}

Deep learning continuously plays a major role in multi-modal systems. It brings strong architectures that can translate one modality into another, improve representation learning, and handle multiple modalities at once \cite{2301.04856}. These methods have been used in many fields, such as healthcare, robotics, video understanding, and text-to-image generation. All of these show that multi-modal learning has great potential in real applications.

However, there are still many open challenges. One of them is how to design more general and powerful models that can handle unseen modalities at inference time \cite{2304.04385}. Another big challenge is how to align, fuse, and reason across heterogeneous data sources in an efficient way \cite{2209.03430}. For large-scale use, models must also be both scalable and computationally efficient.

In recent years, researchers have increasingly focused on the robustness of multi-modal systems. Some studies propose frameworks to test or improve the stability of models under different conditions \cite{2304.04385}. Robustness is essential for making models more reliable and useful in real-world tasks. At the same time, the field needs more shared definitions, taxonomies, and evaluation standards to compare methods to guide future development \cite{1705.09406,2103.06304}.

In summary, multi-modal learning is a dynamic area with the power to change how machines interact with complex data from many sources. If we can solve problems like integration, robustness, and efficiency, for example, by optimizing models such as CLIP for low-resource situations \cite{2301.04856}, we may unlock huge improvements in areas like autonomous driving and medical diagnosis \cite{2209.03430}. In the long run, progress on scalability, generalizability, and new model architectures will be the key to pushing this field forward.

\section{Foundations of Multi-modal Processing}

The foundations of multi-modal processing are rooted in the ability to integrate and handle information from different sources or modalities: text, images, audio, and video, in an effective way. Recent advancements in this field have made significant progress. Various important concepts and techniques have emerged as crucial components of multi-modal machine learning \cite{2209.03430}. One work proposes a taxonomy with six core technical challenges: representation, alignment, reasoning, generation, transference, and quantification \cite{2209.03430}, demonstrating the complexity and wide range of this field.

Representation learning is a cornerstone, aiming to extract shared features across modalities to enable effective integration~\cite{2103.06304}. This process typically involves optimizing a joint loss function that balances modality-specific reconstruction with shared representation constraints.
\begin{equation}
\min_{\theta} \mathcal{L} = \sum_{m=1}^M \mathcal{L}_m(\mathbf{z}, \mathbf{x}_m; \theta_m) + \lambda \mathcal{R}(\mathbf{z}),
\label{eq:shared_representation}
\end{equation}
where $\mathbf{z}$ denotes the shared latent representation, $\mathbf{x}_m$ represents the input data from modality $m$, $\mathcal{L}_m$ is the modality-specific loss, $\mathcal{R}(\mathbf{z})$ enforces regularization (e.g., sparsity), and $\lambda$ balances the trade-off. This formulation ensures the model captures both modality-specific details and cross-modal commonalities.
Alignment, another critical challenge, involves mapping features from different modalities into a shared space. Contrastive learning, as used in models like CLIP, is a common approach to achieve this~\cite{2301.04856}.

\begin{figure}[htbp]
    \centering
    \includegraphics[width=0.5\textwidth]{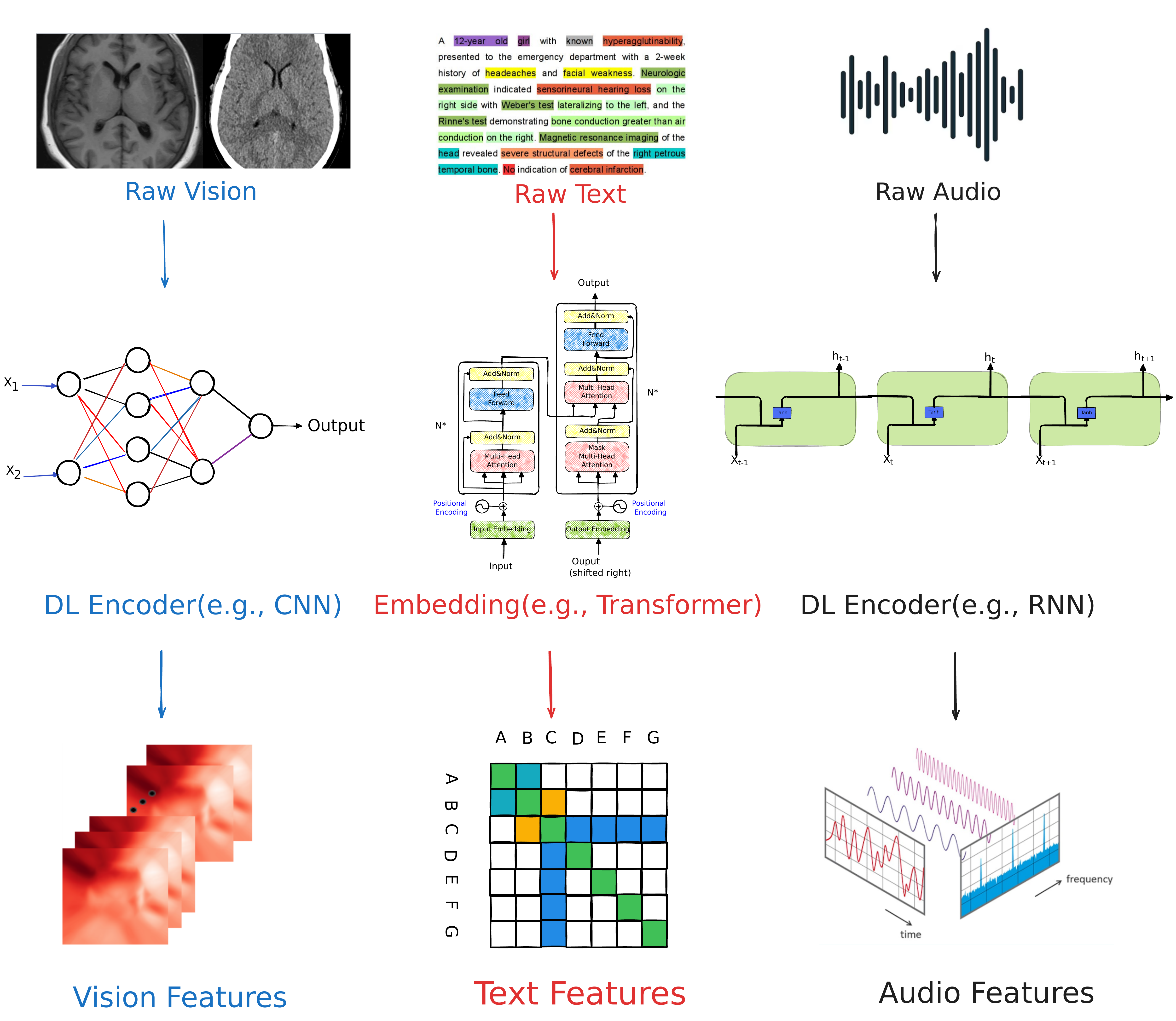}
    \caption{The initial stage of unimodal representation learning, where raw Visual, Text, and Audio inputs are processed by distinct deep learning encoders. Modality-specific architectures transform these diverse inputs into structured Vision Features, Text Features, and Audio Features for subsequent multi-modal integration.}
    \label{fig:multi-modal_representation}
\end{figure}

\begin{equation}
\mathcal{L}_{\text{contrastive}} = -\sum_{i} \log \frac{\exp(\text{sim}(\mathbf{z}_i^1, \mathbf{z}_i^2)/\tau)}{\sum_{j} \exp(\text{sim}(\mathbf{z}_i^1, \mathbf{z}_j^2)/\tau)},
\label{eq:contrastive_loss}
\end{equation}
in this loss function, $\mathbf{z}_i^1$ and $\mathbf{z}_i^2$ are representations of paired modalities (e.g., image and text), $\text{sim}(\cdot, \cdot)$ is a similarity function (e.g., cosine similarity), and $\tau$ is a temperature parameter. This objective maximizes the similarity of matched pairs while minimizing that of unmatched ones.

A holistic evaluation framework is important for understanding the performance of multi-modal foundation models. Such frameworks evaluate models in three dimensions: basic abilities, information flow, and practical use in real-world scenarios \cite{2407.03418}. This highlights the need for better tools to measure how well multi-modal models work, and what they can really do. Meanwhile, some recent surveys have updated how we classify methods. It is no longer just about early vs. late fusion. Many works now focus on more fine-grained strategies for combining modalities \cite{1705.09406}.

One of the major advancements is the use of task-relative definitions of multi-modality. These definitions focus on what kind of information is useful for a specific task \cite{2103.06304}. This view supports more flexible and targeted system designs. Surveys on multi-modal co-learning also address tough problems like missing or noisy modalities. They offer new taxonomies of techniques and applications, which help guide future research \cite{2107.13782}.

Some common themes can be found across these developments. Researchers increasingly see multi-modality as a core aspect of modern machine learning. They also recognize a set of technical and practical challenges, and propose classification systems to better structure the field. The ongoing work on evaluation frameworks, fusion strategies, and co-learning shows that this area is constantly evolving and full of new directions \cite{2301.04856}, \cite{1806.06371}.

There is also growing interest in making multi-modal learning more robust. Some studies offer frameworks to analyze or improve model stability \cite{2304.04385}. At the same time, the potential of multi-modal large language models (MLLMs) continues to attract attention \cite{2311.13165}. The long-term goal is to create generalist multi-modal AI systems that can handle a wide variety of tasks and modalities, an exciting and promising direction \cite{2406.05496}.

However, challenges and limitations remain. Scalability is a major concern. So is the problem of missing or noisy modalities. And there is still no common standard for evaluating multi-modal models \cite{2202.09195}, \cite{2305.03125}. To solve these issues, researchers are exploring methods such as correlation maximization or minimization \cite{2305.03125}, as well as improved data processing techniques for modern multi-modal architectures \cite{2407.19180}.

In conclusion, the foundations of multi-modal processing involve a rich mix of ideas, tools, and useful cases. With continued research, we can expect significant improvements in how we understand and integrate cross-modal data in higher-level models \cite{1806.06371}, \cite{2311.13165}. These advances will influence many areas, like healthcare, education, entertainment, and more: underscoring the importance of continued support and funding for this field.


\section{Deep Learning for Multi-modal Data}

Deep learning has driven major progress in multi-modal research over recent years. Many novel architectures and models have been developed to integrate and process multiple types of data~\cite{2301.04856,2207.02127}. A key direction in this field is multi-modal co-learning, which uses information from one modality to support another, especially when data is missing or noisy~\cite{2107.13782}. Among these, variational autoencoders (VAEs) are often used to create multi-modal generative models that learn shared features~\cite{2207.02127}.

\begin{equation}
\mathcal{L}_{\text{VAE}} = \mathbb{E}_{q_\phi(\mathbf{z}|\mathbf{x})}[\log p_\theta(\mathbf{x}|\mathbf{z})] - D_{\text{KL}}(q_\phi(\mathbf{z}|\mathbf{x}) || p(\mathbf{z})),
\label{eq:vae_elbo}
\end{equation}
here, $q_\phi(\mathbf{z}|\mathbf{x})$ is the encoder’s approximate posterior, $p_\theta(\mathbf{x}|\mathbf{z})$ is the decoder’s likelihood, and $D_{\text{KL}}$ ensures the posterior aligns with a prior $p(\mathbf{z})$. This framework is effective for learning shared representations across modalities.

Architectural design for multi-modal fusion is another major research focus. For instance, EmbraceNet offers a structure that works with different types of models and avoids performance loss when some modalities are missing~\cite{1904.09078}. Cross-modal attention mechanisms are also important, helping align and fuse inputs from different modalities~\cite{1705.09406}. However, scalability and computational cost remain serious concerns~\cite{2006.09310}.

Various deep learning-based methods are widely applied in this field, including neural networks, autoencoders, and generative models \cite{2006.08159}, \cite{2202.09195}. Among these, variational autoencoders (VAEs), generative adversarial networks (GANs) and so on, are often used to create multi-modal generative models that can learn shared features and generate cross-modal outputs \cite{2207.02127}. Cross-modal attention mechanisms are also important. They help align and fuse inputs from different modalities, making integration more effective \cite{1705.09406}.

Even with these advancements, the field still faces several challenges. Scalability and computational cost remain serious concerns \cite{2006.09310}, \cite{2004.12070}. Handling large and diverse datasets can be expensive and complex. In addition, missing or noisy data can harm model performance, and the lack of standard architectures and evaluation benchmarks makes it hard to fairly compare different approaches. These issues show that further research is urgently needed.

Surveys of recent studies reveal several common themes: a strong focus on multi-modal fusion, the use of deep learning frameworks, concerns about robustness, and the importance of organizing the field through taxonomies and benchmarks \cite{2105.11087}, \cite{2211.15837}. Generative models and co-learning strategies continue to attract attention. Still, researchers propose different architectures depending on task type and application needs, leading to varied designs and viewpoints.

Overall, deep learning has brought meaningful progress to multi-modal learning by enabling better integration and processing of diverse data types. However, problems like computational cost, model comparison, and noise handling remain. As the field evolves, we can expect to see more advanced models and better solutions, moving toward more effective, adaptable, and robust multi-modal systems \cite{2212.00101},\cite{2301.04856}.

\section{Multi-modal Fusion Methods}

Fusion methods play a central and often sensitive role in multi-modal learning. They are not just technical add-ons; they are what actually make it possible to bring different modalities: text, vision, audio, or others, together in a way that works. Without good fusion, the promise of multi-modal learning becomes hard to fulfill. That’s why fusion strategies are treated as a core research focus in the community. One insight that has emerged is the importance of learning strong uni-modal features, even under supervised multi-modal training. In particular, \cite{2305.01233} introduces a kind of late-fusion learning method that aims to help models generalize better. What’s interesting here is that their approach captures the finer details unique to each modality, while also trying to limit the harm from noisy or less helpful ones. This is especially useful when the data environment is unpredictable or noisy, which is often the case in real-world settings.

\begin{figure*}[htbp]
    \centering
    \includegraphics[width=\textwidth]{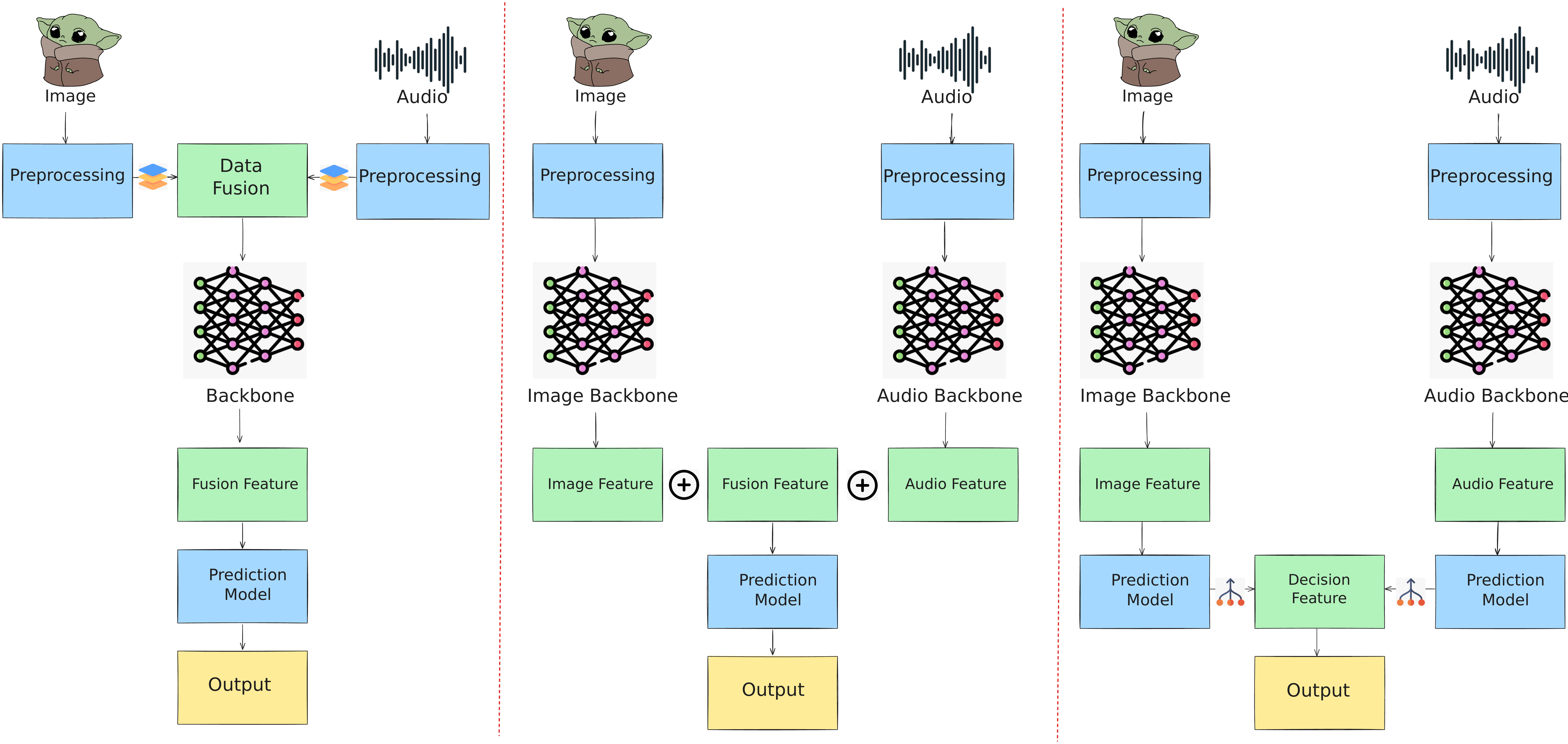}
    \caption{A comparative illustration of three primary multi-modal fusion strategies, Early Fusion, Intermediate Fusion (feature-level), and Late Fusion (decision-level), as applied to image and audio data. Each pipeline depicts a distinct architectural approach for combining information: Early Fusion integrates modalities after initial preprocessing before a shared Backbone; Intermediate Fusion combines features extracted from modality-specific Backbones; and Late Fusion merges the outputs from separate Prediction Models for each modality.}
    \label{fig:multi-modal_fusion}
\end{figure*}

Early fusion combines raw features at the input level, formalized as a weighted combination of modality-specific features~\cite{2301.04856}.

\begin{equation}
\mathbf{h}_{\text{fused}} = f\left( \sum_{m=1}^M \mathbf{W}_m \mathbf{x}_m + \mathbf{b} \right),
\label{eq:early_fusion}
\end{equation}
where $\mathbf{x}_m$ denotes the input features of modality $m$, $\mathbf{W}_m$ is a weight matrix, $\mathbf{b}$ is a bias term, and $f(\cdot)$ is a non-linear activation function (e.g., ReLU). Late fusion, conversely, integrates predictions from modality-specific models, as explored by~\cite{2305.01233}.

\begin{equation}
y_{\text{fused}} = \sum_{m=1}^M w_m \cdot g_m(\mathbf{x}_m; \theta_m),
\label{eq:late_fusion}
\end{equation}
where $g_m(\mathbf{x}_m; \theta_m)$ is the prediction model for modality $m$, $w_m$ is the modality weight, and $y_{\text{fused}}$ is the final prediction. To address modality competition, adaptive gradient modulation dynamically adjusts contributions during training~\cite{2308.07686}.

\begin{equation}
\nabla_{\theta} \mathcal{L} = \sum_{m=1}^M \alpha_m(t) \cdot \nabla_{\theta} \mathcal{L}_m,
\label{eq:adaptive_gradient}
\end{equation}
where $\mathcal{L}_m$ is the loss for modality $m$, $\alpha_m(t)$ is a time-varying weight, and $\nabla_{\theta} \mathcal{L}$ is the fused gradient, ensuring balanced learning across modalities. Intermediate fusion methods, particularly in biomedical applications, allow gradual interaction during training~\cite{2408.02686}.

Other researchers are taking a somewhat different path. Instead of relying purely on late-fusion, they propose models that can learn how different modalities interact and support each other, on their own. For example, the work in \cite{2306.16645} presents a deep equilibrium model for fusion, which has shown quite impressive results on many benchmarks. The model tries to capture high-level dependencies between modalities in a very flexible way. Then, there is \cite{2206.06367}, which argues that there is really no best fusion method that works for all problems. Depending on the task, the modality types, and even how much memory the system can use, the right fusion strategy may be very different.

Somewhere in the middle, intermediate fusion methods offer another option. These are particularly interesting for domains like biomedical applications \cite{2408.02686}, where the way signals mix can be quite subtle and context-sensitive. Intermediate fusion allows for gradual interaction between modalities during training, which often leads to better performance overall. Additionally, \cite{2308.07686} proposes a method using adaptive gradient modulation. This helps reduce what's known as modality competition, where different data sources ``compete'' for influence during learning. They also suggest a new metric to measure this competition, called competition strength.

Looking across these studies, there is a common theme: fusion is not just about mixing things together. It needs to be done carefully, and often in a way that adapts to both the data and the task. Some papers like \cite{2305.01233} argue for late-fusion to keep each modality's strengths intact, while others such as \cite{2408.02686} see value in intermediate fusion, especially for structured or medical data. Adaptive methods like those in \cite{2306.16645} and \cite{2308.07686} focus on the idea that the relationship between modalities can shift, and that the model needs to follow those shifts dynamically. Also, as \cite{2206.06367} mentions, resource availability like memory can shape what fusion strategy is even possible in practice.

An issue that shows up again and again is evaluation. How do we know that fusion is working well? How do we compare models? Multiple works, including \cite{2305.01233} and \cite{2308.07686}, point out the lack of widely agreed-upon metrics. Without proper tools to evaluate fusion quality or modality interaction, it is very hard to say what works and what does not.

And then there is application context. Some methods are proposed for general use \cite{2206.06367}, while others are more specific, like those focused on biomedical data \cite{2408.02686}. This shows the variety in this field but also points to how challenging it is to create fusion methods that are flexible yet powerful across very different tasks and domains.

Even though we’ve seen a lot of progress, there are still problems that have not been solved well. Modality competition, for one, is still a bit of a black box. While \cite{2308.07686} suggests ways to measure and handle it, we do not fully understand the underlying dynamics yet. There is also the point that intermediate fusion methods, which seem promising in medical areas, have not been tested enough in other domains \cite{2408.02686}. So, we do not yet know how far they can go.

To sum up, multi-modal fusion methods are essential, there is no question about that. The field is now focused on finding smarter and more adaptive fusion designs. These aim to balance task-specific performance with computational feasibility. At the same time, better evaluation strategies and metrics are urgently needed. By paying closer attention to how modalities interact, and sometimes interfere, with one another, future models may not only become more accurate but also more transparent and easier to work with. That’s the hope, at least, as multi-modal learning moves ahead.

\begin{table*}[t]
\centering
\caption{Methodological Comparison of Multi-modal Learning Approaches}
\label{tab:method_comparison}
\begin{tabular}{@{}p{2.5cm}p{1.2cm}p{2.8cm}p{2.5cm}p{3cm}p{3.5cm}@{}}
\toprule
\textbf{Method} & \textbf{Year} & \textbf{Fusion Approach} & \textbf{Handles Missing Data} & \textbf{Key Advantage} & \textbf{Primary Application Area} \\
\midrule
EmbraceNet\cite{1904.09078} & 2019 & Modular structure & Yes & Works with different model types & Classification, integration \\
\addlinespace
Late-fusion\cite{2305.01233} & 2023 & Late-fusion & Not specified & Captures finer details of each modality & Generalization in noisy settings \\
\addlinespace
Intermediate fusion\cite{2408.02686} & 2024 & Intermediate & Yes & Gradual interaction during training & Biomedical applications \\
\addlinespace
DEQ-Fusion\cite{2306.16645} & 2023 & Deep & Not specified & Impressive results on benchmarks & General multi-modal tasks \\
\addlinespace
Adaptive Gradient\cite{2308.07686} & 2023 & Adaptive & Not specified & Reduces modality competition & Balanced learning \\
\addlinespace
Greedy Selection\cite{2210.12562} & 2022 & Selection-based & Yes & Computational efficiency & Resource-constrained environments \\
\addlinespace
Teacher-Student\cite{2103.14431} & 2021 & Knowledge transfer & Yes & Works with unlabeled data & Cross-modal learning \\
\addlinespace
Factorized\cite{1806.06176} & 2018 & Factorized & Yes & Focus on shared structures & Incomplete data scenarios \\
\addlinespace
Asymmetric\cite{2501.01240} & 2025 & Asymmetric reinforcement & Yes & Guides model to rely on stable signals & Robust learning \\
\addlinespace
Lightweight Adaptation\cite{2310.03986} & 2023 & Adaptive & Not specified & Resource-efficient & Practical applications \\
\bottomrule
\end{tabular}
\end{table*}

\section{Applications of Multi-modal}

The definition of multi-modality has gone through tremendous change over the past few years, with novel definitions emerging to suit the era of machine learning \cite{2103.06304}. Multi-modal learning has seen great success in applications like computer vision, natural language processing, and speech recognition~\cite{2408.14491}. Recent developments in Multi-modal Large Language Models (MLLMs) have brought improvements to tasks like vision-language pre-training and representation learning~\cite{2210.09263}. New types of data fusion, including early and late fusion, have enhanced efficiency~\cite{2408.14491}, as formalized in~\eqref{eq:early_fusion} and~\eqref{eq:late_fusion}. However, challenges remain, such as the need for better fusion techniques and large-scale datasets~\cite{2412.17759}.

Recent developments in MLLMs have brought dramatic improvements to numerous tasks, including vision-language pre-training \cite{2210.09263} and multi-modal representation learning \cite{2302.00389}. The application of transformer-based models has been particularly successful at reaching state-of-the-art levels in numerous multi-modal tasks \cite{2408.01319}. Still, there are challenges and limitations despite recent advances in multi-modal learning. For instance, the necessity for better fusion techniques to amalgamate data from various modalities is one of the major challenges \cite{2408.14491}. Furthermore, the creation of large-scale data sets that can be used to train and test multi-modal models is still an existing challenge \cite{2412.17759}.

Recent years have seen major advances in both the methodologies and the technical details involved in multi-modal learning. New types of data fusion, including mid-fusion and late-fusion, have enhanced the efficiency of multi-modal models \cite{2408.14491}. Graph-based methods have similarly been applied to strengthen reviews of literature as well as to enhance the analysis of multi-modal data \cite{2408.14491}. In addition, pretraining objectives, including masked language models and next sentence prediction, have been found to be useful to enhance the performance of MLLMs \cite{2408.01319}.

And yet, even with these gains, the future of multi-modal learning remains full of questions, and opportunities. A lot of attention is now focused on designing smarter and more adaptable model architectures that can handle a wide range of modalities smoothly and efficiently \cite{2302.00389}. In addition, improving our understanding of how different modalities interact with each other, especially in real-time or noisy settings, remains a significant challenge \cite{2408.14491}. It is not only about merging signals; it is about understanding their relationships and using that understanding to make better decisions.

\begin{figure}[htbp]
    \centering
    \includegraphics[width=\linewidth]{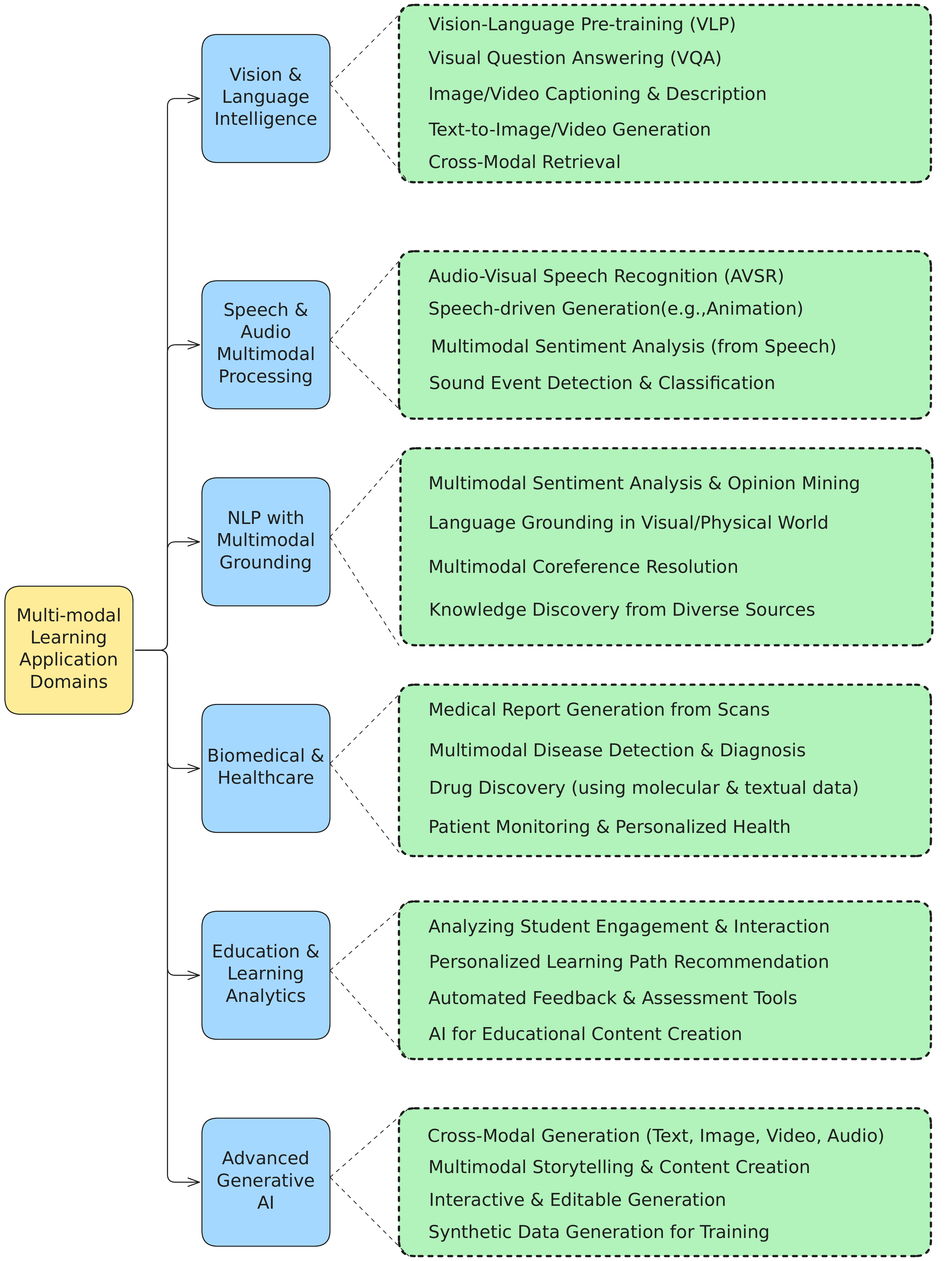}
    \caption{A hierarchical depiction of multi-modal learning application domains, including Vision \& Language Intelligence, Speech \& Audio Processing, NLP with Multimodal Grounding, Biomedical \& Healthcare, Education \& Learning Analytics, and Advanced Generative AI, each exemplified by specific tasks.}
    \label{fig:multi-modal_application}
\end{figure}

Looking forward, we can expect to see even more multi-modal applications in traditional areas like natural language processing, computer vision, and speech recognition. These areas are already mature, but multi-modality brings new possibilities. Ultimately, the long-term goal of multi-modal learning is to create systems that can perform at a level similar to humans on complex, real-world tasks. Systems can see, listen, read, and understand all at once \cite{2412.17759}. That goal may still be far away, but with each new model, dataset, or training method, we get a little bit closer.

\section{Multi-modal Neural Architecture Search}

Multi-modal neural architecture search (NAS) automates the design of efficient neural networks for multi-modal tasks~\cite{2004.12070}. Methods like gradient-based NAS and sequential model-based exploration have achieved state-of-the-art results~\cite{1903.06496}. However, challenges such as computational cost and generalizability persist~\cite{2409.07825}. Fusion strategies, including early and deep fusion, are critical to NAS performance~\cite{2310.17737}.

One predominant theme that is found throughout the literature is the automation of architectural design, aiming to save manual efforts as well as to increase performance \cite{2004.12070}\cite{1903.06496}. Most methods try to offer flexible frameworks that can be capitalized upon across various multi-modal tasks as well as data sets \cite{2104.09379}\cite{2405.17927}. The efficiency of the search process is another target for several researchers to minimize the computational cost as well as the time consumed by NAS \cite{1906.05226}\cite{2003.01181}.

Although these progresses have been made, there exist differing opinions and methods within this domain. To be specific, \cite{1903.06496} is interested in this novel search space exclusively for fusion architectures, whereas others, such as \cite{2004.12070} and \cite{2104.09379} investigate more comprehensive frameworks that support multiple different operations and tasks. The modalities' integration is even addressed differently between studies, where some prioritize early fusion \cite{2301.04856}, while others propose deep fusion or the fusion of both \cite{2310.17737}. In addition, the application of pretrained models differs, with some methods utilizing pretrained unimodal backbones \cite{2104.09379}, where other methods rely on training from scratch or alternative pretraining methods such as \cite{2310.17737}'s Masked Architecture Modeling (MAM).

The technical methodologies employed in multi-modal NAS are diverse and include gradient-based NAS \cite{2004.12070}, sequential model-based exploration \cite{1903.06496}, bilevel searching schemes \cite{2104.09379}, and bi-modal learning \cite{2310.17737}. These approaches have been applied to various multi-modal tasks, including those involving vision, language, and audio modalities \cite{1705.09406}\cite{2302.00389}. However, despite these advancements, challenges persist, including the computational cost associated with NAS \cite{2409.07825}, ensuring generalizability across tasks \cite{2209.03430}, interpreting and explaining the performance of searched architectures \cite{2304.04385}, and scalability to new modalities \cite{2405.17730}.

In summary, multi-modal neural architecture search has emerged as a dynamic and fast-developing field with great potential to improve both the efficiency and effectiveness of multi-modal learning systems. While a number of recent contributions have made meaningful steps toward addressing issues such as cost, transferability, and architectural complexity, much remains to be explored. Questions related to scalability, interpretability, and generalization are still open and continue to limit the full realization of NAS in practical multi-modal applications \cite{2104.09379}\cite{2405.17927}. Looking forward, it is reasonable to expect that more advanced frameworks and algorithms will be introduced to meet these challenges. With further work, we may see models that are not only better suited to handling the unique difficulties of multi-modal learning, but also more capable of delivering consistent performance across a wide array of real-world tasks \cite{2310.17737}\cite{2405.17927}.

\section{Real-Time and Low-Resource Scenarios}
Efficient multi-modal processing has long been a central concern in multi-modal learning, especially under real-time and low-resource conditions. Researchers have focused on integrating information from different modalities in ways that are both accurate and highly efficient. For instance, the survey paper ``Multi-modal Co-learning: Challenges, Applications with Datasets, Recent Advances and Future Directions''~\cite{2107.13782} offers a comprehensive review of multi-modal co-learning, highlighting issues like noisy or missing modalities and proposing a taxonomy to structure the problem. Robustness is essential in such scenarios, and attribution regularization encourages models to utilize all available modalities, mitigating single-modality dominance~\cite{2404.02359}.

\begin{equation}
\mathcal{L}_{\text{total}} = \mathcal{L}_{\text{task}} + \lambda \sum_{m=1}^M \left\| \frac{\partial \mathcal{L}_{\text{task}}}{\partial \mathbf{x}_m} \right\|_2^2,
\label{eq:attribution_regularization}
\end{equation}
where $\mathcal{L}_{\text{task}}$ is the task-specific loss, $\frac{\partial \mathcal{L}_{\text{task}}}{\partial \mathbf{x}_m}$ measures modality $m$’s gradient contribution, and $\lambda$ controls regularization strength. Additionally, greedy modality selection improves efficiency under computational limits by prioritizing informative modalities~\cite{2210.12562}. These approaches reflect the field’s emphasis on clear definitions, structured analysis, and practical challenges.

Another important contribution comes from the paper ``Greedy Modality Selection via Approximate Submodular Maximization'' \cite{2210.12562}. It introduces a state-of-the-art approach with an efficient algorithm designed to select the most informative and complementary modalities under computational limits. The method shows strong results not only on synthetic tasks but also in real-world datasets, making a clear case for modality optimization as a practical way to improve processing efficiency. By contrast, the work titled ``Modular and Parameter-Efficient Multi-modal Fusion with Prompting'' \cite{2203.08055} proposes a lightweight solution using modal prompts and novel vector alignment. The technique performs comparably to other mainstream methods even under low-resource scenarios. This shows that prompting, though relatively new in this field, can be a flexible and scalable approach for multi-modal learning when resources are limited.

Robustness is another essential topic that has gained attention in recent work. One representative study, ``On Robustness in Multi-modal Learning'' \cite{2304.04385}, proposes a framework to identify the weaknesses in commonly used representation learning techniques. It further suggests targeted interventions to improve robustness. In a related direction, the paper ``Attribution Regularization for Multi-modal Paradigms'' \cite{2404.02359} proposes a new regularization term. This term encourages models to better utilize all available modalities, which also helps mitigate the problem of single-modality dominance. These contributions suggest that improving generalizability and robustness is not only possible but necessary, especially as models are applied to real-world environments where data sources may be uncertain or unstable.

A common thread running through these contributions is the growing recognition of efficiency and scalability as core needs in multi-modal learning. For example, the work ``Multi-modal Information Bottleneck: Learning Minimalist Models'' \cite{mai2022multimodal} emphasizes the value of minimalist modeling, how to reduce redundancy while still maintaining performance. This also connects to broader questions around evaluation: what counts as ``good enough'' performance in constrained settings, and how should it be measured across different tasks and systems?

That said, many difficulties remain. Although recent progress is meaningful, challenges in developing advanced modeling techniques still persist. Scalability and real-time efficiency remain two of the most critical bottlenecks, especially in domains where fast decisions are essential. Also, as models begin to deal with increasingly diverse modalities and environments, robustness becomes even more crucial. In addition, a lack of consistent benchmarks and shared evaluation metrics continues to limit how we compare methods and track progress.

Experimental work in this area has offered valuable insight into the tradeoffs involved in integrating multiple modalities under limited resources. The results of recent studies go well beyond academia and point toward wide applications, ranging from healthcare and education to autonomous systems and human-computer interaction. As the field continues to develop, it will be essential to focus not only on overcoming current problems but also on setting clearer standards. That means developing shared evaluation frameworks, refining fusion and selection methods, and continuing efforts to improve robustness and generalizability in a broad range of scenarios. With that, we may move closer to achieving multi-modal processing techniques that are not only efficient and accurate, but also practical and widely usable.

\section{AutoML for Multi-modal Learning: Recent Developments and Future Directions}

The integration of automated machine learning (AutoML) with multi-modal learning has advanced rapidly, driven by increasing data complexity and the demand for efficient, user-friendly solutions~\cite{2408.00665,2404.16233}. Frameworks like AutoM3L, leveraging large language models, automate multi-modal training pipelines, reducing manual effort~\cite{2408.00665}, while AutoGluon-multi-modal (AutoMM) enables fine-tuning of foundation models with minimal code, supporting diverse tasks~\cite{2404.16233}. Simple techniques, such as cross-modal alignment, yield robust results~\cite{2412.16243}, as shown in~\eqref{eq:contrastive_loss}. However, challenges like pipeline complexity and inconsistent evaluation metrics persist~\cite{2111.02705}.

The combination of AutoML with computer vision has also been widely studied. Because data is getting bigger and more complex, researchers try to make AutoML easier to use. AutoM3L \cite{2408.00665} provides an LLM-based solution to help users set up multi-modal pipelines without much effort. AutoMM \cite{2404.16233} also gives a simple and efficient way to handle different modalities and improve performance on many real-world tasks.

Recent studies tell us something very clear: handling multi-modal data well is no longer a bonus, it is a must. The paper ``Bag of Tricks for Multi-modal AutoML'' \cite{2412.16243} runs a large benchmark that covers images, texts, and tables. It shows that even simple tricks, such as fusion, data augmentation, and cross-modal alignment, can bring strong and stable results. But this paper also reminds us that mixing multiple data types is not an easy job. So we need AutoML tools that can link different modalities together quickly and smartly.

Another big trend is the use of advanced models, especially large language models (LLMs) and foundation models \cite{2408.00665}\cite{2404.16233}. These models make it possible to automate very complicated tasks, improve overall results, and let more people use machine learning, even if they are not experts. The shift from rule-based tools to model-based methods shows that the community now trusts LLMs to do reasoning, generate code, and provide useful interactions.

However, even with this exciting progress, there are still many problems. First, since we are dealing with many types of data, the pipelines become heavier and slower. Also, it is still hard to make the models explainable \cite{1907.08392}\cite{1810.13306}. Another issue is evaluation. Right now, different studies use different metrics and datasets. So we cannot compare fairly. One common benchmark is badly needed to help build strong and general AutoML systems \cite{2111.02705}.

Looking back at how AutoML has changed, we can see a big shift. The paper ``Automated Machine Learning , A Brief Review'' \cite{2008.08516} explains this clearly. In the early years, AutoML focused on tuning hyperparameters or picking models. But today, researchers care more about how to use AutoML in real life, especially with messy and mixed multi-modal data. People want tools that are not only fast but also easy to use and able to handle real-world problems.

To sum up, when AutoML meets multi-modal learning, things get very interesting. We now have many strong tools and smart frameworks, but we also face a lot of challenges behind the scenes. If we want to unlock the full power of AutoML, we still need to solve key pain points, like pipeline complexity, scaling up, explainability, and fair benchmarking. Fixing these will make ML tools faster, more friendly, and strong enough to handle difficult tasks. Also, this research is closely linked to the big direction of AI, since more and more LLMs and foundation models are being developed \cite{2408.00665}\cite{2404.16233}. So we believe continuous research and teamwork are the key to pushing AutoML forward in the multi-modal world.

\section{Unsupervised and Semi-Supervised Multi-modal Learning}

Unsupervised and semi-supervised learning are among the fastest-growing directions in multi-modal research, driven by their ability to handle unseen modality combinations at inference time~\cite{2306.12795}. Recent work emphasizes task-specific modality selection, prioritizing utility over comprehensive modeling~\cite{2103.06304}. Theoretical bounds further quantify cross-modal interactions in semi-supervised scenarios~\cite{2306.04539}.

\begin{equation}
\begin{split}
I(\mathbf{X}_1; \mathbf{X}_2 \mid \mathbf{Y}) \leq \min(H(\mathbf{X}_1),\\
H(\mathbf{X}_2)) - H(\mathbf{Y} \mid \mathbf{X}_1, \mathbf{X}_2),
\label{eq:interaction_bound}
\end{split}
\end{equation}
where $I(\mathbf{X}_1; \mathbf{X}_2 \mid \mathbf{Y})$ is the conditional mutual information, $H(\cdot)$ denotes entropy, and $H(\mathbf{Y} \mid \mathbf{X}_1, \mathbf{X}_2)$ is the conditional entropy. Teacher-student frameworks also enhance generalization without labeled data~\cite{2103.14431}.

At the same time, there is an ongoing effort to understand whether different modalities are genuinely interacting, or just existing side by side. Liang et al. \cite{2306.04539} proposed a set of theoretical bounds to estimate the interaction level in semi-supervised scenarios. Sure, these bounds do not offer precise yes-or-no answers, but they do help guide us in both evaluating and designing better systems. Meanwhile, another interesting work \cite{2103.14431} introduced a teacher-student framework where a pre-trained single-modal teacher can still pass meaningful knowledge to a multi-modal student, even in the absence of labeled data. Surprisingly, this approach improves generalization and lowers the model’s sensitivity to noisy signals.

If there is one key message across these studies, it is this: flexibility and robustness are no longer optional, they are essential. Today’s models need to be ready for missing data, noisy inputs, or strange, previously unseen modality mixes. What makes this possible? Strong representation learning. Many teams are now using self-supervised or unlabeled data to learn meaningful features; two good examples can be found in \cite{2301.04856}\cite{2408.14491}. These works push the field forward not just in terms of technical depth, but also in how we understand multi-modal systems in real-world or even complex environmental contexts.

However, not everyone agrees on what ``multi-modality'' should mean. Some papers, like \cite{2103.06304}, advocate for task-specific definitions, meaning the way we define and handle modalities depends on what the task actually needs. Others hold on to more traditional, general-purpose views. Even worse, there is no unified standard for measuring cross-modal interactions. One paper turns to information theory \cite{2306.04539}; others prefer to rely on practical benchmarks or novel metrics. So depending on the lens you use, the evaluation result may vary dramatically.

Technically, the field is vibrant with innovation. For example, \cite{2306.12795} explores novel mixes of projection and fusion. Liang et al. \cite{2306.04539} give mathematical tools, bounds and ranges, to quantify how well different modalities actually interact. \cite{2103.14431} makes the case that a model can still learn without labels, using cross-modal knowledge transfer. On top of that, others like \cite{2110.11601}\cite{2501.01240} are testing semi-supervised setups in tasks like 3D object recognition, or working on ways to reduce biases introduced by imbalanced modality signals.

Of course, many problems are still far from solved. One major pain point is data, large labeled datasets remain costly and time-consuming to collect \cite{1705.09406}\cite{2107.13782}. Another issue is, we still do not have a reliable way to estimate cross-modal interaction without supervision \cite{2306.04539}. And let’s be honest: we are still waiting for that one model that can truly handle any level of supervision and still be robust in the face of messy, real-world input.

To wrap things up, unsupervised and semi-supervised multi-modal learning is definitely moving fast, and it has already delivered some meaningful progress, like learning from unlabeled data, mixing unseen modalities, and estimating interaction without full supervision. But still, we are only scratching the surface. More research is clearly needed to build truly adaptable, flexible, and strong systems. If we succeed, fields like computer vision, natural language processing, and human-computer interaction could benefit from smarter, more resilient models that do not rely so heavily on fully labeled data.

\section{Evaluation Metrics and Benchmarks}
Evaluating multi-modal models is challenging due to messy and diverse data~\cite{2107.07502}. Benchmarks like MultiBench and MM-BigBench test generalization and robustness~\cite{2107.07502,2310.09036}. Missing modalities require diagnostic and intervention methods~\cite{2409.07825}, often using fusion strategies like~\eqref{eq:early_fusion}. Standardized metrics and comprehensive benchmarks remain essential to ensure fair comparisons across tasks and domains.

That’s why recent studies have begun offering us some new tools. One highlight is MultiBench \cite{2107.07502}. This is not just any benchmark, it is a large-scale, unified setup built to test representation learning across 15 datasets, 10 modalities, 20 prediction tasks, and 6 different application areas. Sounds intense? That’s the point. It helps us see how well a model can generalize, how efficient it is in terms of time and memory, and how robust it remains when one modality, say, vision or text, suddenly gets weak or disappears altogether. These tests do not just give scores, they reveal true strength.

Another landmark is MM-BigBench \cite{2310.09036}, which takes a wider angle. It builds a kind of testbed for checking how models handle multi-modal content-understanding tasks. And it does not just look at accuracy. It brings in a bunch of metrics: Best Performance, Mean Relative Gain, Stability, Adaptability, a full 360-degree view. The message is pretty clear: one single number is not enough anymore. We need metrics in bunches, because a multi-modal system is complex and multi-skilled. You miss the full picture if you just look at the top-line score. Adding to that, a review from \cite{2409.18142} surveys 211 benchmarks covering MLLMs across four core abilities, understanding, reasoning, generation, and application. Again, the call is for a clear, shared rulebook. Without that, we are just throwing darts in the dark.

\begin{figure}[htbp]
    \centering
    \includegraphics[width=\linewidth]{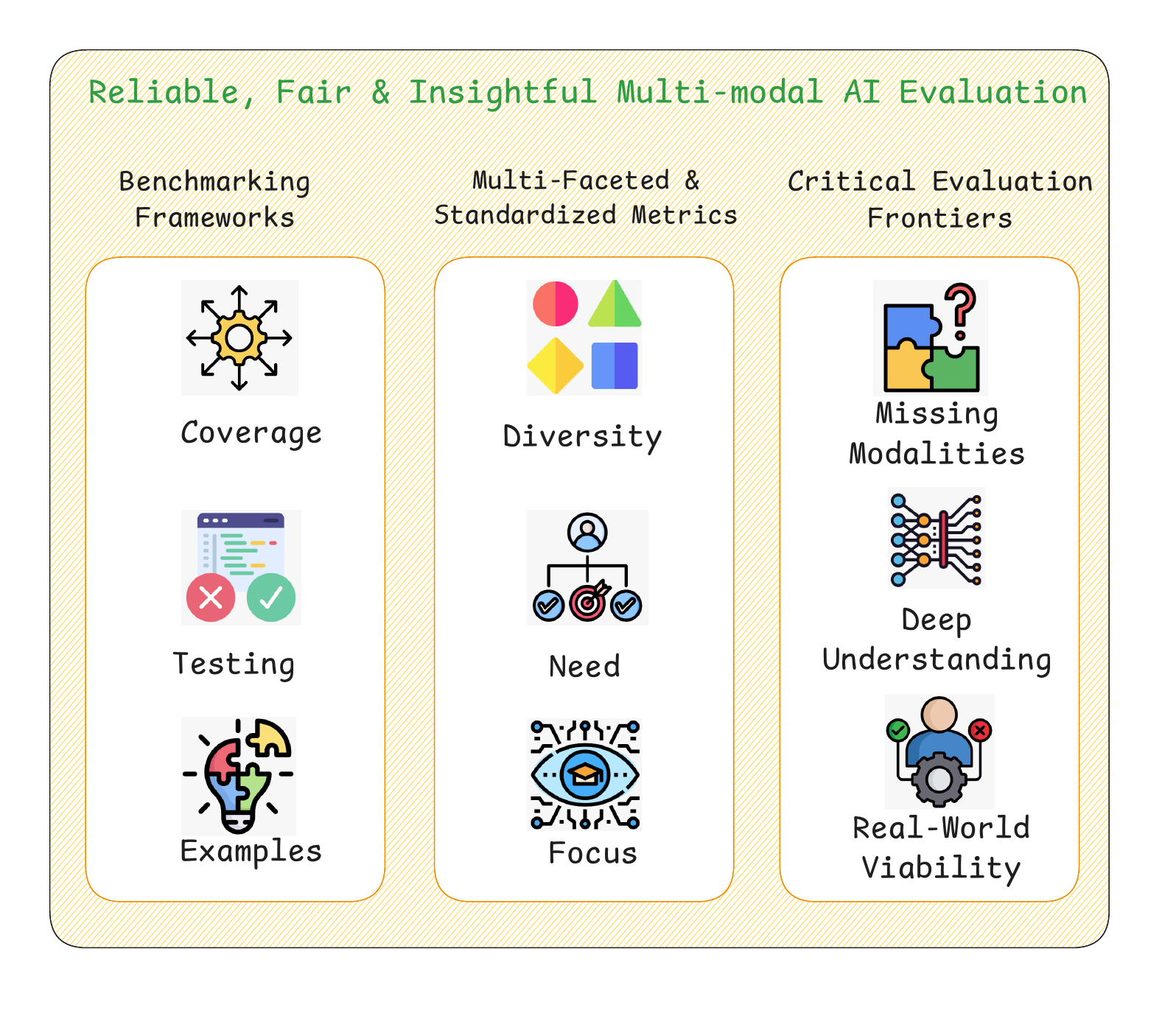}
    \caption{Pillars supporting effective multi-modal evaluation, resting on a foundation of realistic and diverse data. Key components include comprehensive Benchmarking Frameworks (assessing coverage, testing, with examples like MultiBench), Multi-Faceted \& Standardized Metrics (emphasizing diversity and the need for standardization), and addressing Critical Evaluation Frontiers (such as missing modalities, deep understanding, and real-world viability), all contributing to reliable, fair, and insightful multi-modal AI assessment.}
    \label{fig:multi-modal_evaluation}
\end{figure}

Another hot topic? Missing modalities. This is not a rare problem, it happens all the time. Real-world data? Always broken in some way. Maybe a sensor fails. Maybe half the input is corrupted. Papers like \cite{2409.07825}\cite{2304.04385} zoom in on this. They explore both diagnostic tools, how to detect when a modality is gone, and intervention methods, which are tricks to make the system stay strong anyway. These studies remind us: if your multi-modal model only works when all inputs are perfect, then it is not really ready for real life.

Now, if you step back and look across all this literature, a loud voice echoes: we need better benchmarks, and we need better metrics. Public datasets are crucial, not just for transparency, but for actual science. We want repeatable experiments. We want fair, apples-to-apples comparisons between models. That’s why works like \cite{2311.02692}\cite{2408.08632} argue we must go deeper, not just check surface-level matching, but really dig into cross-modal understanding. Does the model just align text to image, or does it actually ``get'' the connection?

Still, not everyone agrees. Some papers prefer single-metric focus, they swear by Best Performance or Mean Relative Gain. Others go for the toolbox approach, mixing lots of indicators. The same split exists when it comes to fixing missing modalities. Some works try simple tricks like imputation. Others go all-in with customized robustness strategies.

From a technical point of view, most of the models are still deep nets, CNNs, RNNs, and especially Transformers. Researchers also explore the usual suspects: early fusion, late fusion, and middle fusion, all tested in works like \cite{2107.07502}\cite{2310.09036}. When it comes to testing stability, some follow the robustness-checking protocols from \cite{2304.04385}, where they purposely inject noise or drop out parts of the modality data and watch how the models behave.

But let’s be real, big problems are still there. Building massive benchmarks takes a ton of compute and money. That makes it hard to scale. Also, without a shared set of metrics, it is tough to compare two models fairly. What looks ``better'' in one paper might actually be weaker in another setting. And worst of all, models trained in clean lab settings often crash when taken out into the wild, where inputs are incomplete, modalities are missing, or data shifts in unpredictable ways.

So, in conclusion: yes, there has been progress. The community now understands that we have to move toward larger, richer, more diverse benchmarks. But we are not done. There are still open problems, like setting standardized scores, designing smarter ways to handle missing modalities, and building truly robust systems. Future work will need to connect all these dots, so we can finally build multi-modal models that not only shine in the lab but also survive and thrive in the messy, chaotic, real-world settings they are meant to serve.

\section{Heterogeneous Data, Missing Modalities, and Adversarial Attacks}

Heterogeneous data and missing modalities pose significant challenges in multi-modal learning~\cite{2409.07825}. Factorized representations focus on shared structures when data is incomplete~\cite{1806.06176}.

\begin{equation}
\mathbf{z} = \sum_{m \in \mathcal{M}_{\text{avail}}} \mathbf{W}_m \mathbf{x}_m + \mathbf{b},
\label{eq:factorized_representation}
\end{equation}
where $\mathcal{M}_{\text{avail}}$ is the set of available modalities, $\mathbf{W}_m$ is the projection matrix, and $\mathbf{z}$ is the robust representation. Asymmetric reinforcement prioritizes stable modalities~\cite{2501.01240}.

\begin{equation}
\mathcal{L}_{\text{asym}} = \sum_{m \in \mathcal{M}_{\text{avail}}} w_m \mathcal{L}_m + \gamma \sum_{m \notin \mathcal{M}_{\text{avail}}} \mathcal{L}_{\text{impute}}(\mathbf{x}_m),
\label{eq:asymmetric_reinforcement}
\end{equation}
in this equation, $\mathcal{L}_m$ is the loss for modality $m$, $w_m$ is its weight, and $\mathcal{L}_{\text{impute}}$ is the imputation loss. Adversarial attacks require robust defenses~\cite{1810.08437}.

\begin{equation}
\mathcal{L}_{\text{adv}} = \max_{\|\delta\|_p \leq \epsilon} \mathcal{L}(f(\mathbf{x} + \delta), y),
\label{eq:adversarial_loss}
\end{equation}
where $\delta$ is a perturbation with $p$-norm bound $\|\delta\|_p \leq \epsilon$, $f(\cdot)$ is the model, and $\mathcal{L}(\cdot, y)$ is the task loss.

Over the past few years, multi-modal learning has made impressive progress. We’ve seen strong performance in a wide range of domains, computer vision, natural language processing, healthcare, and more \cite{2409.07825}\cite{1806.06176}. But still, let’s be honest: this field is far from ``solved.'' Several deep-rooted challenges continue to limit how far we can go. Some of them are technical; others are more about how the real world works. Either way, we need to face them head-on.

One of the biggest headaches is dealing with heterogeneous data, that is, information coming from totally different sources, each with its own format and structure \cite{2112.12792}. Imagine a dataset combining text, images, and audio. Each of these modalities has its own rhythm, its own kind of noise, its own ``language.'' And to expect a single model to handle them all smoothly? It is hard. The model must learn not just from each modality but also across them, and that’s where things start to break down.

Another major challenge? Missing modalities. And trust me, this is not rare. A sensor might fail. Data gets corrupted. Or, in some cases, certain signals are deliberately excluded for privacy or cost reasons \cite{2409.07825}\cite{2103.05677}. The result? A performance drop, sometimes a big one. So now, the question becomes: how do we build models that still function when part of the input goes missing? Some recent solutions offer promising ideas. For example, factorized representations help models focus on shared structures even when data is incomplete \cite{1806.06176}. Others propose asymmetric reinforcement \cite{2501.01240}, guiding the model to rely more on stable signals. And then there are lightweight adaptation techniques \cite{2310.03986}, which offer practical, less resource-intensive fixes. These methods show one thing clearly: resilience is now a must.

And resilience is not just about noise or gaps, it is also about adversarial attacks \cite{1810.08437}. In high-stakes environments like medical AI or autonomous driving, even tiny, human-imperceptible changes to input data can lead to completely wrong predictions. That’s scary. So, defending against such attacks, detecting them, resisting them, recovering from them, has become a core concern \cite{2411.14717}. Multi-modal models need to be not just smart, but safe and secure.

Then there is the issue of modality imbalance, a subtle but serious problem. Sometimes one modality dominates the learning process because its signals are stronger or easier to model. This leads to the model ``ignoring'' weaker but still-important signals \cite{2210.12562}\cite{2112.12792}. And that’s bad, because it reduces the diversity of information the model actually uses. So, what do we do? Some researchers suggest greedy modality selection \cite{2210.12562}, where the model picks the most useful signals step by step. Others turn to game-theoretic regularization \cite{2411.07335}, where learning is balanced like a strategic game between modalities. Both methods aim to reduce bias and boost fairness in training, something increasingly critical in today’s AI landscape.

Looking across all this work, one thing is obvious: solving these core issues is key to building stronger, more reliable multi-modal systems \cite{2409.07825}\cite{2304.04385}. It is not just about improving one metric here or there. It is about improving how we learn joint representations, how we deal with data loss, and how we protect against manipulation. And interestingly, many of the proposed techniques are cross-cutting. For example, factorized learning and asymmetric reinforcement are useful not just for missing data, but also for handling imbalance and improving robustness overall \cite{1806.06176}\cite{2501.01240}. That kind of versatility is powerful.

In summary, while the field of multi-modal learning has achieved notable progress, a range of persistent challenges continues to limit its broader adoption and effectiveness. Key technical obstacles, such as the integration of heterogeneous data, resilience to missing or corrupted modalities, and defense against adversarial interference, remain unsolved. Existing approaches, including factorized learning frameworks, robust training techniques, and lightweight adaptation methods, have demonstrated value across multiple problem domains rather than in isolation. This suggests that future research should focus on developing integrated solutions capable of addressing overlapping challenges within a unified framework. By understanding the interconnections between these problems, and by designing models that reflect such complexity, the field can move closer to producing truly robust, adaptive, and generalizable multi-modal systems suitable for deployment in real-world, imperfect environments \cite{2409.07825}\cite{2304.04385}.

\section{Future Directions in Multi-modal Research}

As multi-modal learning continues to grow, clearer frameworks and practical taxonomies are increasingly critical~\cite{2408.14491,2209.03430}. These tools help researchers manage the complexity of heterogeneous data and ambiguous integration rules~\cite{1705.09406,2107.13782}. Structuring the field is as vital as advancing model design.

A key direction is the push for task-relative definitions of multi-modality, focusing on modalities relevant to specific tasks~\cite{2103.06304}. This approach enhances efficiency by prioritizing utility over exhaustive modeling, though challenges arise with unbalanced or misaligned modalities~\cite{2411.17040,2406.05496}. Deep learning for co-learning and robust fusion strategies, such as~\eqref{eq:early_fusion} and~\eqref{eq:late_fusion}, are gaining traction~\cite{2301.04856}, enabling modalities to guide each other in challenging settings~\cite{2311.13165}.

Despite progress, challenges persist. Combining diverse data types is inherently complex, with issues like unstructured, noisy, or missing data~\cite{2409.07825}. Large-scale labeled datasets, particularly for non-text modalities, remain scarce~\cite{2107.07502,2405.19334}. These gaps hinder training flexible and reliable models, necessitating adaptive systems for imperfect conditions.

The field’s interdisciplinary nature, blending computer science, cognitive research, and beyond, will drive future growth. Flexible, real-world-ready solutions are essential, balancing theoretical advances with practical constraints like limited resources and incomplete labels. Ultimately, improving definitions, fusion strategies, and evaluation tools without overcomplicating systems will unlock multi-modal learning’s potential in AI, healthcare, and education~\cite{2408.14491}.

\section{The Evolving Landscape of Multi-modal Learning}

The field of multi-modal learning has gone through noticeable changes in recent years. This shift has been largely driven by the rapid progress in areas like machine learning, computer vision, and natural language processing. According to studies such as \cite{2408.14491}\cite{2103.06304}\cite{1705.09406}\cite{2209.03430}\cite{2301.04856}, the field has grown into a dynamic research hotspot, with a wide range of tasks and challenges emerging along the way. A shared belief among many researchers is that combining different types of data, text, image, audio, or beyond, can offer a more complete and nuanced understanding of complex problems \cite{2107.13782}\cite{2312.06037}. This idea has inspired the creation of new definitions, conceptual models, and frameworks to address the unique challenges that arise when working with multi-modal inputs \cite{2103.06304}\cite{2309.12458}.

Across the literature, some recurring patterns begin to surface. A common one is the difficulty of handling multiple modalities: not just processing them in isolation, but figuring out how to connect them, align them, and model their interactions in a meaningful way \cite{1705.09406}\cite{2209.03430}. At the same time, more and more researchers are turning to deep learning as a backbone for these efforts, applying it across domains such as healthcare, education, and autonomous systems \cite{2301.04856}\cite{2304.04385}. In parallel, identifying open questions and mapping future directions has become a key feature of many recent surveys and reviews \cite{2209.03430}\cite{2412.17759}, a sign that the field is still very much evolving.

Interestingly, there is no complete consensus on what multi-modality even means. Some researchers propose their own definitions, some build entirely new taxonomies, and others focus on different fusion strategies, early, late, mid-level, or even graph-based approaches. For instance, \cite{2408.14491} provides a comprehensive literature review, while \cite{2103.06304} introduces a redefined concept of modality interaction. These variations do not contradict each other so much as reflect the broad and multifaceted nature of the field.

From a technical standpoint, the methods explored in these papers are also quite diverse. Some focus on theory and modeling, while others are more application-oriented. For example, \cite{2110.14202} investigates meta-learning through a multi-task lens, and \cite{2305.03125} proposes a strategy that alternates between increasing and decreasing inter-modal correlations to improve learning outcomes. All of this suggests that multi-modal learning is not a single discipline but a meeting point of many others, requiring interdisciplinary knowledge and flexibility in thinking.

Still, despite all the enthusiasm, several core difficulties remain. There are calls for deeper theoretical work to solidify the foundations of this area \cite{2209.03430}\cite{2309.12458}. Mixing and managing different types of data continues to be a technical bottleneck \cite{1705.09406}\cite{2301.04856}. Moreover, training large multi-modal models often demands significant computational resources and access to large-scale, high-quality datasets \cite{2301.04856}\cite{2412.17759}. Another concern that comes up again and again is the lack of transparency: many current models are still hard to interpret or explain in practice \cite{2209.03430}\cite{2312.06037}, which limits trust and usability in real-world settings.

Overall, multi-modal learning has come a long way, and it still has a long way to go. Thanks to developments in machine learning and related fields, the tools we now have are more powerful and versatile than ever. But alongside this progress are real challenges, both technical and conceptual. What makes this area especially exciting is its potential: the promise of building smarter, more robust, and more interpretable models. With continued research and refinement, we can expect to see new breakthroughs and broader applications, especially in areas like computer vision, natural language processing, and human-computer interaction \cite{2107.13782}\cite{2312.06037}\cite{2408.14491}.

\section{References and Further Reading}

The field of multi-modal learning has been growing rapidly, with a wide range of studies exploring its many directions and subfields. This section provides a brief overview of some representative works, highlighting shared challenges, technical advances, and common lines of thought. For instance, deep learning–based models have now reached state-of-the-art performance in tasks such as Visual Question Answering (VQA), Natural Language for Visual Reasoning (NLVR), and Vision-Language Retrieval (VLR), as reported in \cite{2301.04856}. A comprehensive review from \cite{2408.14491} also shows how integrating multiple modalities can offer a deeper view into learning behaviors and performance patterns.

Among the most frequently mentioned concerns is the issue of missing or noisy data, as well as the high cost of collecting labeled samples, problems that appear in multiple studies like \cite{1804.08010}\cite{2107.13782}. One promising response has been the development of multi-modal co-learning. For example, \cite{2107.13782} presents a detailed summary of how co-learning works, what it has achieved so far, and where it might go next. Similarly, \cite{2302.00389} offers an extensive survey on the evolution of multi-modal representation learning, tracing the development of deep architectures, training objectives, and pretraining methods across the years.

Transformer-based models now dominate much of this space. Both \cite{2301.04856}\cite{2302.00389} demonstrate how these architectures help push performance further across tasks in computer vision, natural language processing, and education. This trend is reinforced by other works like \cite{2408.14491}\cite{1804.08010}\cite{2107.13782}, which show just how widely multi-modal methods are being applied across different domains.

Some papers focus on targeted use cases, for example, co-reference resolution, discussed in \cite{1804.08010}, while others step back to offer higher-level syntheses, laying out the broader evolution of the field and where it is headed next \cite{2408.14491}\cite{2302.00389}\cite{2107.13782}. A number of studies also compare supervised and semi-supervised learning settings. For example, \cite{1804.08010} advocates for semi-supervised approaches when labeled data is scarce, especially in low-resource environments.

From a technical angle, the toolbox is diverse. Researchers use CNNs, RNNs, and Transformers \cite{2301.04856}\cite{2302.00389}, often in combination with transfer learning, multi-task setups, and co-learning strategies \cite{2107.13782}. Attention mechanisms, particularly self-attention and cross-attention, are frequently integrated to strengthen modality alignment and interaction \cite{2301.04856}\cite{2302.00389}.

Still, several issues remain unresolved. Annotated datasets are still missing for many modality types \cite{1804.08010}\cite{2107.13782}, and data quality, especially when modalities are weak or noisy, remains a bottleneck \cite{2107.13782}. Seamless fusion is another challenge: many models struggle to make different modalities work together effectively \cite{2301.04856}\cite{2302.00389}. Additionally, as highlighted in \cite{2408.14491}\cite{2302.00389}, the field still lacks strong, unified benchmarking standards, making it difficult to compare models fairly.

Beyond these main threads, other valuable studies include \cite{2201.01234}, which focuses on vision-based multi-modal applications, and \cite{2105.06701}, which proposes a novel fusion approach. Meanwhile, \cite{2007.06326} digs into some of the field’s deeper conceptual and technical difficulties, while also offering directions for future research.

All in all, the literature shows a field that is active, experimental, and still developing. While many exciting models and methods have emerged, much remains to be done. These works offer a strong base, but also point toward the many open problems that will define the next chapter of multi-modal learning.

\section{Conclusion}

Multi-modal learning is no longer an emerging field, it is steadily becoming a foundational direction in machine learning, driven by the growing need to handle data that is rich, diverse, and often incomplete. From dynamic fusion strategies to modular architectures, and from contrastive alignment to self-supervised pretraining, the community has introduced a remarkable array of technical solutions. Many of these models now demonstrate impressive performance across benchmarks in vision-language reasoning, retrieval, and multi-modal generation. Yet as the field grows more sophisticated, so too do its tensions. There remains a fundamental gap between what current architectures can optimize in closed settings, and what real-world systems demand in terms of robustness, interpretability, and resource awareness. In many ways, recent progress has shifted the difficulty from representation to integration: knowing how to combine modalities is no longer enough; we must now learn when and why certain combinations work, or fail.

At the same time, evaluation remains an underdeveloped frontier. Most existing metrics still fall short of capturing cross-modal interaction quality, modality imbalance, or robustness under noise and missing inputs. This misalignment between model capability and evaluation standards has led to increasing calls for richer, task-aware benchmarks that reflect practical constraints, rather than isolated accuracy scores. Parallel to this, ethical concerns, about bias propagation, opacity in decision-making, and the cost of large-scale multi-modal training, are entering the conversation more directly. These concerns cannot be deferred to deployment stages; they must be built into model design and benchmarking from the beginning.

Looking ahead, the future of multi-modal learning may lie not in making models ever larger or more universal, but in making them more adaptive, more explainable, and more integrated into human-centered workflows. There is growing interest in architectures that support modality dropout, resource-aware computation, and on-the-fly adaptation, especially in areas such as education, assistive technology, or edge AI. Progress in this field will likely require tighter collaboration between machine learning researchers, domain experts, and cognitive scientists, not only to build systems that perform well but also to ensure they align with how humans interpret, combine, and prioritize information in complex environments. As infrastructure improves, and as interdisciplinary dialogue deepens, multi-modal learning holds the promise of reshaping how we model knowledge, not just across data types, but across disciplines, values, and contexts.

\bibliographystyle{apalike}
\bibliography{cite}

\end{document}